\documentclass[journal]{IEEEtran}
\usepackage{siunitx}
\usepackage{amssymb}

\usepackage{graphicx}
\ifCLASSINFOpdf
\else
\fi
\begin{document}
%
\title{Leveraging the Potential of Novel Data in Power Line Communication of Electricity Grids}
%
%
%

\author{Christoph~Balada,~Sheraz~Ahmed,~Andreas~Dengel,~Max~Bondorf~and~Markus~Zdrallek
\thanks{C. Balada, S. Ahmed and A. Dengel are with the German Research Center for Artificial Intelligence (DFKI), Kaiserslautern, Germany; \{christoph.balada, sheraz.ahmed, andreas.dengel\}@dfki.de}
\thanks{M. Bondorf and M. Zdrallek are with University of Wuppertal, Wuppertal, Germany;\{bondorf, zdrallek\}@uni-wuppertal.de}}

%
%

\markboth{IEEE Transactions on Smart Grid,~Vol.~XX, No.~XX, August~20XX}%
{C. Balada \MakeLowercase{\textit{et al.}}:Novel Data in Power Line Communication of Electricity Grids}
%



\maketitle

\begin{abstract}
Electricity grids have become an essential part of daily life, even if they are often not noticed in everyday life.
We usually only become particularly aware of this dependence by the time the electricity grid is no longer available. 
However, significant changes, such as the transition to renewable energy (photovoltaic, wind turbines, etc.) and an increasing number of energy consumers with complex load profiles (electric vehicles, home battery systems, etc.), pose new challenges for the electricity grid.
At the same time, these challenges are usually too complex to be solved with traditional approaches. In this gap, where traditional approaches are reaching their limits, Machine Learning has become a popular tool to bridge this shortcoming through data-driven approaches. 
Machine learning (ML) is a source of untapped potential for a variety of different challenges that the electricity grid will face in the future if we had enough data.
To enable novel ML implementations is we propose FiN-2 dataset, the first large-scale real-world broadband powerline communications (PLC) dataset.
Encouraged by the pioneering efforts that went into compiling the FiN-1 dataset, we address a limitation of FiN-1, namely the small size and spatial coverage of the data. 
FiN-2 was collected during real practical use in a part of the German low-voltage grid that supplies energy to over 4.4 million people and shows well over 13 billion data points collected by more than 5100 sensors.
In addition, we present different use cases in asset management, grid state visualization, forecasting, predictive maintenance, and novelty detection to highlight the benefits of these types of data.
For these applications, we particularly highlight the use of novel machine learning architectures to extract rich information from real-world data that cannot be captured using traditional approaches. 
By publishing the first large-scale real-world dataset, we also aim to shed light on the previously largely unrecognized potential of PLC data and emphasize machine-learning-based research in low-voltage distribution networks by presenting a variety of different use cases.
\end{abstract}


\begin{figure*}[h]
    \includegraphics[width=\textwidth,trim={0cm 0cm 0cm 0cm},clip]{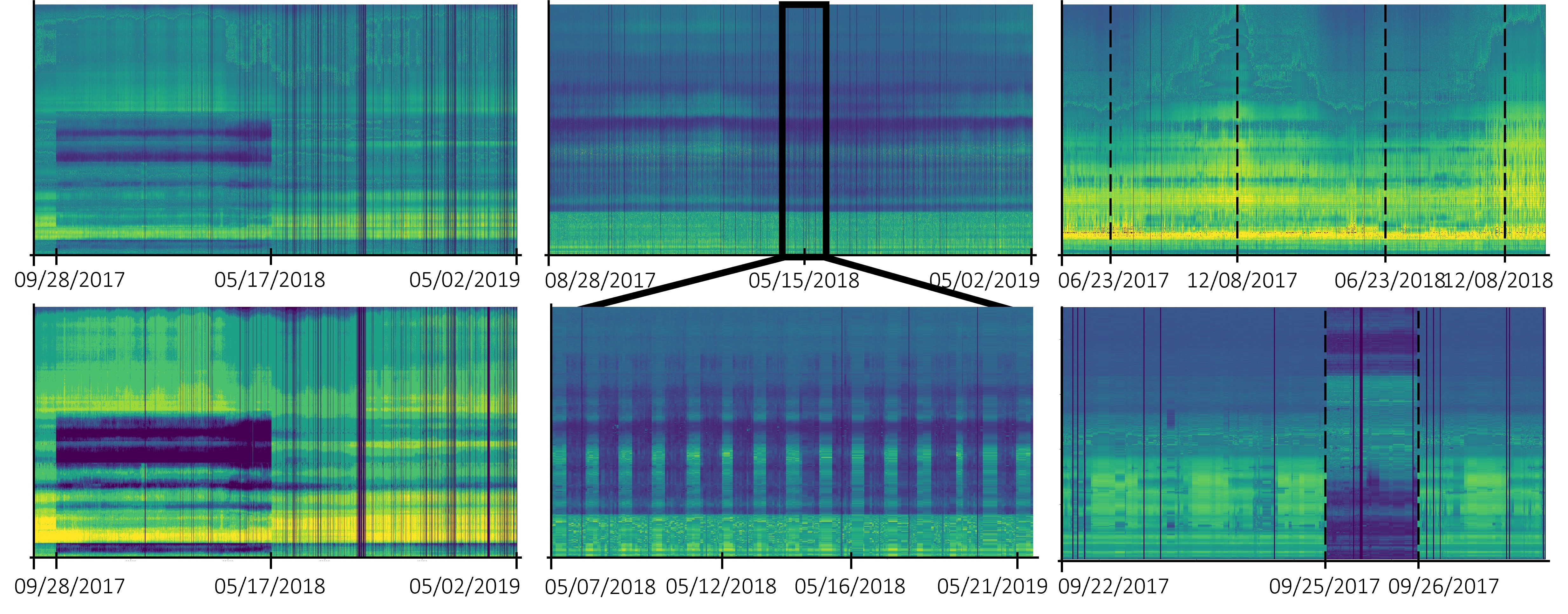}
    \caption{Different examples of SNR spectra. The left figures show an example of an SNR spectrum (top) with the corresponding tone map (bottom). The middle figures show an example of a recurring noise (top two years; bottom two weeks). The right figure shows a slight noise at the top, which recurs seasonally, and the failure of a fuse at the bottom.}
    \label{fig:data_overview}
\end{figure*}

%
\IEEEpeerreviewmaketitle
\section{Introduction}\label{sec:intro}
Electricity grids play an indispensable role in providing the basic infrastructure for our modern lives.
At the same time, electricity grids are currently facing one of the most significant changes since their invention. On the one hand, we have the transition to renewable energies to combat climate change, and on the other hand, there’s a rise in digitisation and automation \cite{galli2011grid, miller2015status} efforts.
While the need to change the way energy is generated, transmitted and distributed is obvious from a climate change perspective, the implicit impacts and opportunities of the transition to a smart grid are far more subtle.
In this context, Power Line Communication (PLC) has proven to be a reliable foundation for providing a communication infrastructure capable of interconnecting smart grid components such as smart meters \cite{galli2011grid, lampe2011power, bavarian2012physical, dostert2000fundamental}. 
However, in addition to providing a communication infrastructure, PLC can do even more for future smart grid systems \cite{miller2015status, international2018status,wruk2021optimized}.
PLC nodes form a distributed measurement network that is able to provide near real-time insights into large parts of the network, and especially into the low-voltage part of the grid.
Particularly for low-voltage grids (lv grid), this new possibility of monitoring represents a breakthrough \cite{miller2015status}.
However, the actual value of this data is still largely unrecognized.

While these challenges posed by the transition to a smart grid are typically too complex to be solved by traditional approaches, Machine Learning (ML) has become a popular tool to bridge this shortcoming through data-driven approaches. 
Machine Learning is a source of untapped potential for a variety of different challenges that the electricity grid will face in the future if we had enough data. Inspired by the need for more data to implement such novel ML approaches and encouraged by the pioneering efforts that went into compiling the FiN-1 dataset, we present the FiN-2 dataset, which reshapes the standards for publicly available data in this area.

In this paper, we showcase the novel FiN-2 dataset and highlight the benefits of data-driven approaches based on data from FiN-1 and FiN-2 dataset. To enable research in these application areas, we are releasing the FiN-2 dataset, addressing the immense data hunger of modern ML that paves the way to use cases ranging from asset management, grid state visualization, forecasting, predictive maintenance to novelty detection. To underline these use cases, we show several practical examples from different application areas and emphasize the advantages for network operators and other stakeholders. 

Inspired by that, we address a limitation of FiN1, namely the low spatial extent and number of sensors and measurements in the PLC network.
FiN-2 dataset was collected during real practical use in a part of the German lv grid that supplies energy to over 4.4 million people and shows more than 13 billion data points collected by more than 5100 sensors.
We pay particular attention to the signal-to-noise ratio (SNR) measurements between the PLC nodes, and to the voltage measurements at the nodes.
Our main contribution is two-fold: first, we present the first large-scale real-world dataset of PLC-based data, paving the way towards more practical smart grid research based on real-world data. Secondly, we give a broad overview of how ML-based PLC analysis can provide added value in domains like grid monitoring, asset management or anomaly detection.

\section{Overview of the proposed FiN datasets}
The proposed novel FiN-2 dataset as well as FiN-1 dataset \cite{balada2022fin}  were collected in the context of two successive projects, and consist of SNR spectra of PLC connections complemented by various measurements such as the nodes' voltage under operation. 
Figure \ref{fig:data_overview} shows an SNR spectrum example for a 1-to-1 PLC connection within the lv grid.
Two types of PLC modems were involved in the process of data generation. 
Data in FiN-1 was solely collected by PLC-repeater-modems installed in local substations and cable distribution cabinets. FiN-2 also includes data from smart meter gateways, which are installed in households.
In general, single SNR values indicate the quality of a connection on a single PLC channel, in a single time step. Therefore, the SNR spectrum can be interpreted as the connection quality over the entire bandwidth of the channels over time.
The connection quality is influenced by a wide range of different factors including potential interactions between grid components such as fuses, cables, cable joints, street lightning or sources of interference like bad shielded electronics.

FiN-1 as well as FiN-2 dataset hold primarily data from PLC nodes that were distributed in the lv grid. However, FiN-1 also contains some nodes connected to the grid on medium-voltage (mv) level.
While FiN-1 covers only 68 PLC connections, FiN-2 involves about 24,000 connections.
Nevertheless, both datasets have their respective advantages and focus on different use cases.
FiN-2 is well suited for investigating effects affecting large areas of the electricity grid due to the large footprint of the numerous nodes and FiN-1, with its wider range of metadata, is better suited for investigating individual connections. 
Table \ref{table:data_overview} summarizes the key characteristics of both datasets.

Another aspect is the location of the PLC nodes. 
As FiN-1 was a pilot project, all nodes are located in three areas that show typical urban and suburban surroundings. 
While FiN-2 covers most of the nodes installed during FiN-1, the number of nodes significantly increased. 
Therefore, FiN-2 shows a wide range of different environments such as urban, suburban, rural, industrial, downtown, small villages, etc.  
FiN-2 locations connect around 4.4 million customers.
The following sections present an outline of both datasets regarding the data that is published.  

\begin{table}[b]
    \centering
    \begin{tabular}{l|ll}
         \hspace{2cm} &   FiN-1    &    FiN-2  \\ 
    \multicolumn{3}{l}{Base data} \\ \hline
    \multicolumn{1}{l|}{SNR measurements} & 3.7 Million & $>$ 1 Billion \\
    \multicolumn{1}{l|}{Voltage, thd, phase measurements} & - & $>$ 12 Billion \\
    \multicolumn{1}{l|}{Time span} & 2.5yrs & 1.5yrs \\
    \multicolumn{1}{l|}{Distinct PLC connections} & 68 & 23937 \\
    \multicolumn{3}{l}{ } \\
    \multicolumn{3}{l}{Metadata} \\ \hline
    \multicolumn{1}{l|}{Cable joints} & \checkmark & \checkmark* \\
    \multicolumn{1}{l|}{Cable properties (length, type, etc.)} & \checkmark &  \\
    \multicolumn{1}{l|}{Year of installation} & \checkmark &  \\
    \multicolumn{1}{l|}{Weather} & \checkmark & \checkmark* \\
    \end{tabular}
    \caption{key characteristics of the FiN-1 and FiN-2 dataset. \\ * in a limited scope}
    \label{table:data_overview}
\end{table}

\begin{figure*}[h!]
    \includegraphics[width=\textwidth,trim={0cm 0cm 0cm 0cm},clip]{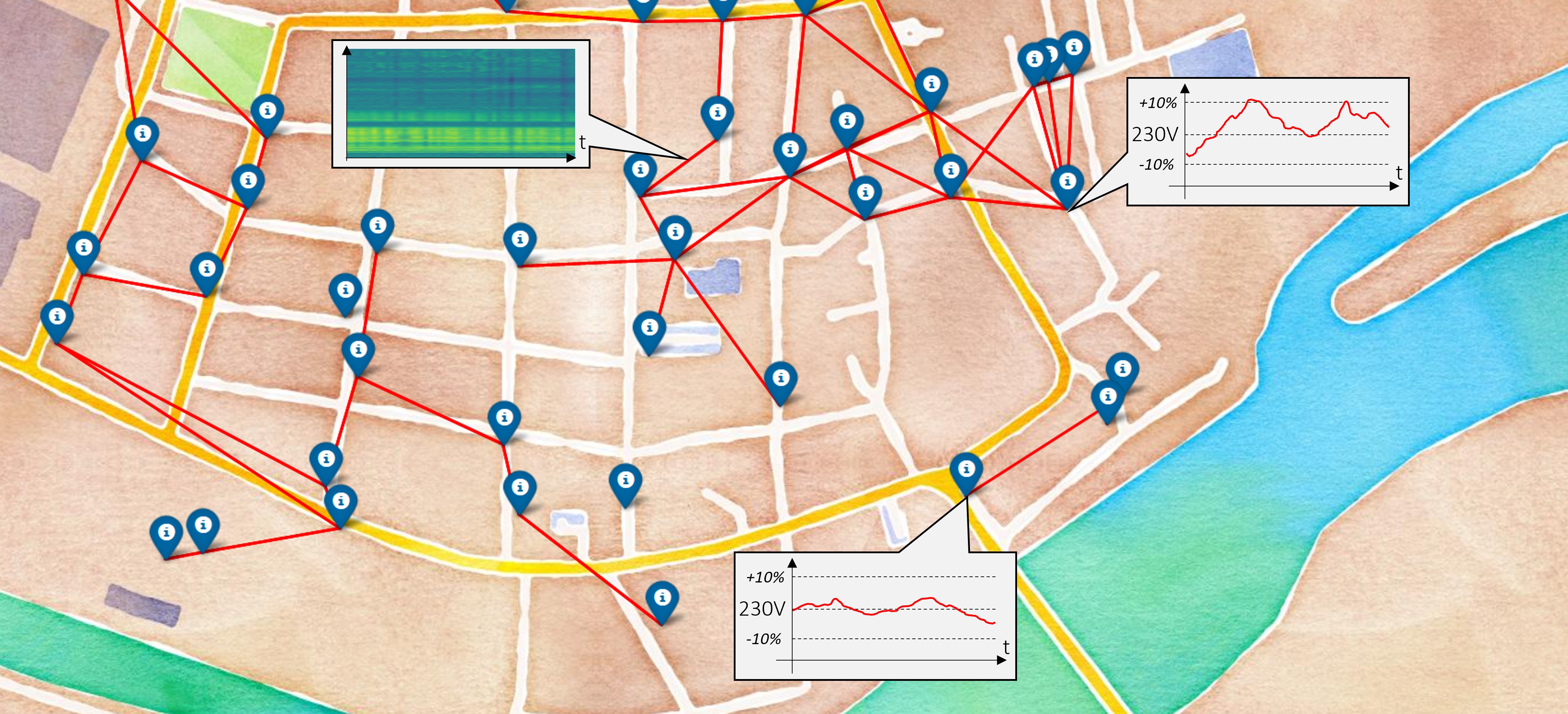}
    \caption{Example of a lv grid monitoring system. PLC technology opens up new possibilities for monitoring wide areas of the electricity grid at lv-level. A map-based interface can provide insights into grid states and conditions. In addition to visualizing measured values such as SNR or voltage, the map can also be used to display results from downstream ML applications such as anomaly detection, estimated cable life, or detected fuse failures. The map shows an actual PLC installation from the FiN-2 project.}
    \label{fig:grid_monitoring}
\end{figure*}

\subsection{FiN-1}
Over a period of about 2 years, 68 1-to-1 connections and their SNR spectra were measured for FiN-1. 
A total of 3.7 million SNR measurements were performed in an interval of 15 min and in addition, the measurements are supplemented by a wide range of metadata. 
This includes cable related metadata such as age, type, number of conductors, length, cross-section, number of cable joints and environment related metadata such as temperature, rain, sun hours and cloudiness.
Each PLC connection uses a total of 917 different channels, with the SNR measured for each channel individually.
The SNR is specified in a range from \SI{0}{dB} to \SI{40}{dB}.
In addition, the tonemap was also saved, which indicates in eight different levels which modulation was used for the PLC transmission. 
Ultimately, this results in a set of three vectors for each PLC connection in FiN-1: the SNR spectrum (of shape $t \times 917$ in $\SI{}{dB}$), the tonemap (also of shape $t \times 917$) and a corresponding list of timestamps, of shape $t \times 1$, where $t$ is the number of measurements in a 15-min interval.
Figure \ref{fig:data_overview} shows an example of an SNR spectrum and a tonemap.

\subsection{FiN-2}
Over a period of about 2 years, more than 5100 nodes, multiple SNR spectra, as well as other measured values, were recorded for FiN-2. 
In addition to the SNR spectrum, voltage, total harmonic distortion and phase angle were also recorded for each of the three phases. 
Some of the metadata collected for FiN-1 is also available for FiN-2, but it was only possible to collect this metadata for a limited number of nodes.
The measurement of the SNR spectrum, as well as the tonemap, was performed in analogy to the procedure from FiN-1, in a 15-min interval. 
Unlike FiN-1, the use of better measuring electronics allowed the SNR spectrum to be determined in a range of \SI{-10}{dB} and \SI{40}{dB}.
In total, the FiN-2 system comprises about 29 million SNR spectra and about one billion nodal measurements (voltage, etc.) per month.

\section{Challenges in using PLC data and fields of application}
Data were recorded in various German cities and were collected during practical use in the field. 
Due to the enormous number of PLC nodes, several terabytes of data were generated. 
Processing this enormous amount of data is a challenge for which current ML approaches are ideally suited and even benefit.
Compared to laboratory experiments or simulations, there is an immense number of stationary and dynamic sources of interference compared to practical operation, as well as events that have an influence on the SNR spectrum. 
It is common that the results of laboratory experiments are difficult to apply in real-world applications due to this complex set of influences. 
Furthermore, fluctuations and trends occur on the basis of hours, days or even seasonally. 
Due to all these influences, as well as the individual cable properties, each connection has an individual characteristic in the SNR spectrum \cite{ferreira2011power}.
Depending on the type of task, be it classification, regression or forecasting, it is a challenge to extract the intended information from the SNR spectrum. 
Frequently, the noise components that affect the SNR signal are considerably larger than information about special events and trends in the intended signal.
An example of this would be the estimation of the remaining lifetime of a cable based on the SNR spectrum \cite{hopfer2020nutzen,forstel2017grid}.
A change in cable characteristics due to ageing, on the one hand, has an impact on the SNR spectrum, but on the other hand, it is a process with very slow and small effects so that they are usually exceeded by orders of magnitude by daily or seasonal trends.
Apart from these numerical challenges, there are also very practical problems that arise when continuous real-time processing of such data is required.
The proposed FiN datasets represent a unique opportunity to analyse peculiarities occurring in the operation of lv electricity grids, develop strategies to overcome potential challenges and explore new application fields of data from PLC networks.
In the following, we have a look at the bigger picture and list just a few possible areas of application.

\begin{figure}[h!]
    \centering
    \includegraphics[width=0.49\textwidth,trim={0.05cm 0cm 0cm 0cm},clip]{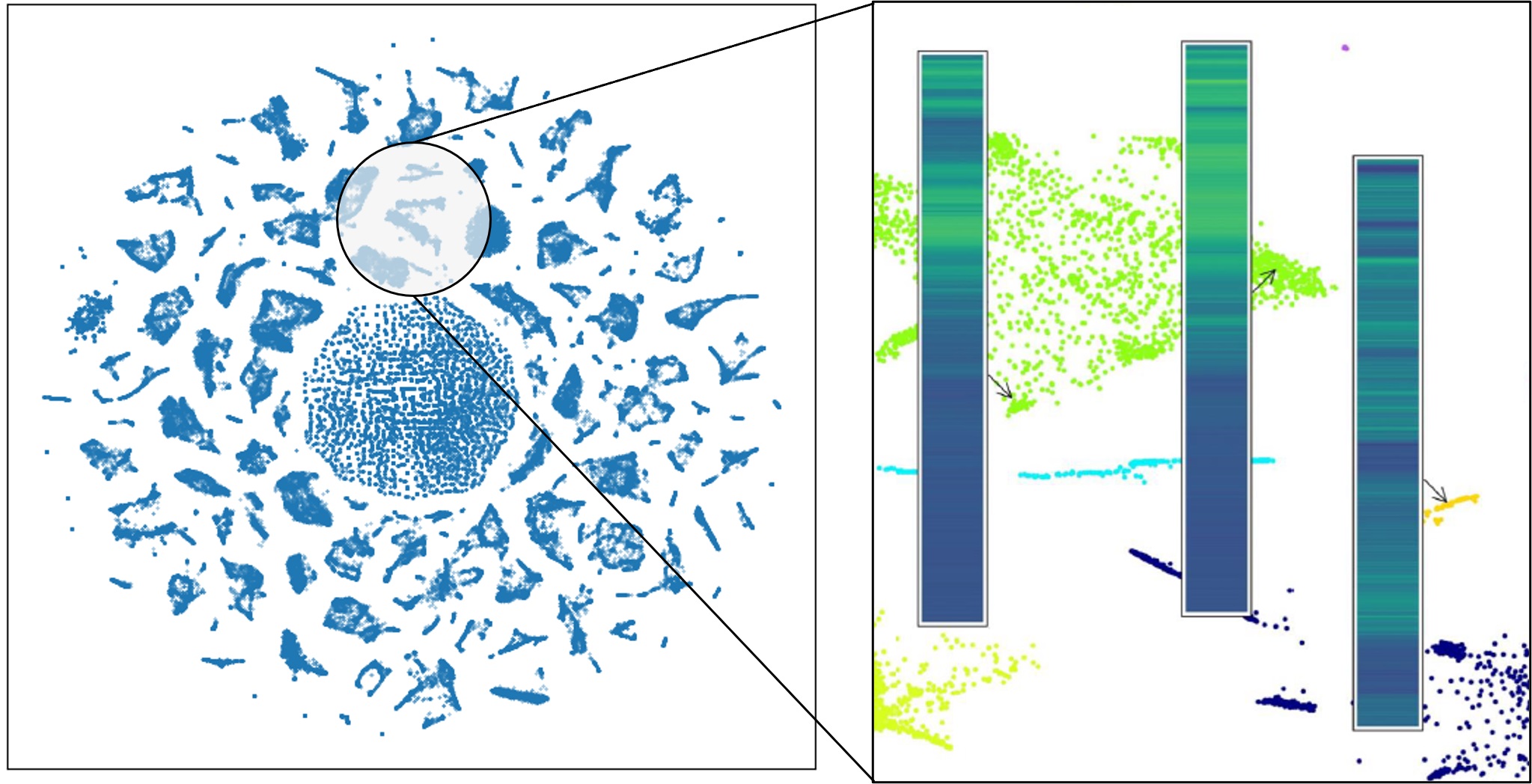}
    \caption{t-SNE visualization of a FiN-1 subset. The left plot shows the overall landscape of the different clusters, while the right plot shows a zoomed-in view with different SNR samples. Whereas samples within the same cluster show similar SNR profiles, samples from other clusters show different characteristics.}
    \label{fig:cluster_w_sample}
\end{figure}

\subsection{Fields of application}\label{sec:use_cases}
To meet future climate change requirements, it is necessary to abandon fossil fuels and move to renewable energy sources, such as photovoltaic (PV) systems.
However, this will inevitably lead to the emergence of more complex load profiles due to, for instance, the prevalence of electric transport vehicles \& PV installations, which have an increasing impact on power system planning \cite{international2018status}.
Traditional grid control is thus increasingly reaching its limits.
Therefore, having a distributed sensor network, such as a PLC network for grid automation, becomes critical. For example, in \cite{koch2021voltage}, the use of PLC modems as voltage sensors for smart grids and grid automation was investigated.
With the help of voltage data of the FiN-2 dataset, we enable further research in this area. In particular, for fields like grid monitoring, asset management and anomaly detection.

\textbf{Grid Monitoring:} Whereas the grid is already extensively monitored at mv and hv-level, PLC is now opening up the possibility of introducing comprehensive monitoring at lv level as well \cite{hopfer2020nutzen}.
Potential problems at lv level include: Fuse, cable and joint failures which result in power outages. While the power outages themselves can be detected directly through area-wide monitoring, the other aspects can be addressed by processing the SNR spectrum.

\textbf{Asset Monitoring:} Research in SNR spectra from PLC systems shows that fuse failures can be detected with a simple SNR analysis \cite{bondorf2021broadband}.
In addition, laboratory tests have indicated that SNR can be used to detect the condition of a cable \cite{hopfer2019analysis}.
Ultimately, live monitoring of grid assets such as power cables will be part of future smart grid systems.

\textbf{Grid Condition:} In addition to the assessment of grid assets, the wide range of effects that influence the SNR spectrum opens up new possibilities for analysing the condition of the grid.
\cite{huo2021power} shows how SNR forecasting could be used to discover anomalies in the PLC network. 

\textbf{Security \& anomaly detection:} Especially if PLC expansion continues, the security aspect is of growing importance. 
On the one hand, this concerns the reliability of PLC communication to ensure communication with smart grid components such as smart meters or customers. 
On the other hand, it also concerns security against attacks on control systems of the electricity grid. 
While anomaly detection could be incorporated to detect and locate interference sources that disturb PLC connections \cite{bondorf2021broadband}, voltage measurements of PLC systems that are independent of legacy grid control systems, could be used to verify the grid state.
Since PLC modems form a distributed network whose nodes measure and communicate independently, PLC networks are comparatively hard to compromise. 
As the grid state is subject to a local continuity, irregular measurements can be validated by neighbouring nodes. 
Neighbourhoods in the PLC network can therefore be checked against each other for plausibility. 

\section{Example uses of FiN dataset}
As highlighted in \ref{sec:use_cases} we want to underline the novelty of these datasets. 
This concerns, on the one hand, the large amount of data from an extensive collection of different PLC nodes, as well as the fact that all data comes from actual use during normal grid operation.
In the following, we want to support the aforementioned fields of application with explicit examples. 
Thereby, we see ML as an important tool in future grid research.

\subsection{Live monitoring of the lv grid}
The comprehensive monitoring made possible by PLC will be key to addressing the challenges of transitioning to a future smart grid system.

This applies in particular to the lv grid, which can be monitored over a large area for the first time through the use of PLC.
Figure \ref{fig:grid_monitoring} shows a pseudonimized example of how a PLC network in the lv grid could be visualised. 
Using maps facilitates manual inspection and underpins interpretability for humans. 
Furthermore, a map can be used to track down network issues, to find bottlenecks in communication infrastructure or to report about grid assets.
Live reporting of the nodes enables an immediate response to network issues like overvoltage, cable breaks, fuse failures etc.

However, such a visualisation is just a frontend for grid operators and could be used as a foundation to visualize findings from downstream ML models like the following example use cases. 
To demonstrate the novelty of the data, we present in the following various ML applications that fall within the context of monitoring and asset management.
All applications are based solely on the SNR data and are performed during normal operation of the electricity grid.

\begin{figure*}[h!]
    \centering
    \includegraphics[width=\textwidth,trim={0cm 0cm 3.6cm 0cm},clip]{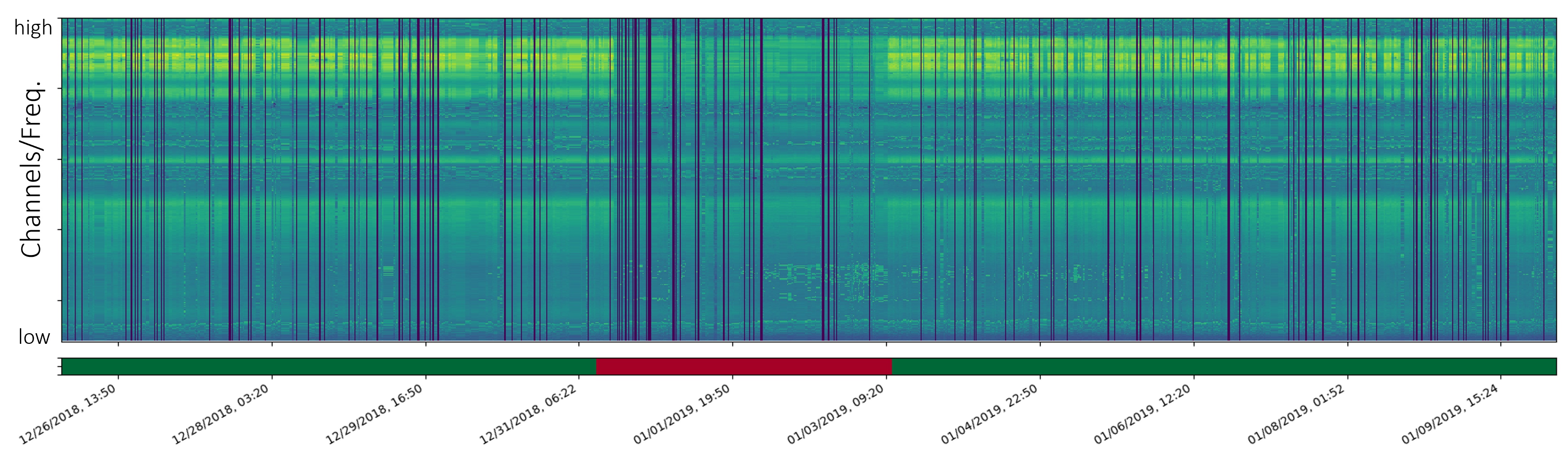}
    \caption{SNR Spectrum of about 2 weeks with an anomaly indicator provided by DTW. The spectrum shows an SNR section of about 3 days that were highlighted as anomaly due to a breakdown that especially affects the high-frequency channels. Anomalous sections are marked in red and normal sections in green.}
    \label{fig:dtw}
\end{figure*}

\subsection{Connection states}\label{sec:node_state}
While general grid monitoring concerns the grid as a whole, the SNR assessment also allows monitoring of the state of a single connection.
SNR spectra contain rich information regarding cable properties, trends, noise sources, fuse failures and anomalies. 
To exploit this hidden information, we propose an ML approach, that uses embeddings and clustering to identify connection states.   

In order to create such an embedding and enable its visualization, we propose, an embedding that maps the data to two dimensions.
This process consists of two steps, one of which is optional.
The first step creates a distribution of data points that divides data points into semantically similar clusters.
Semantic similarity, in this application, refers to the similarity of the SNR profile.
In a second step, the resulting high-dimensional data distribution is projected into a low-dimensional space (in our case, 2 dimensions). 
Since initial experiments indicated that the data had sufficient semantic structure without further processing, we omitted the first step and used the raw SNR data directly for dimensionality reduction.
While a traditional principal component analysis (PCA \cite{pearson1901liii}) requires eight dimensions to capture at least 95\% of the variance in the FiN-2 dataset, we propose t-SNE \cite{van2008visualizing} as an ML approach to compute a low-dimensional representation. 
The aim of t-SNE is to capture the topology of the high-dimensional data in a low-dimensional space. 
In addition to t-SNE, we also tested UMAP \cite{mcinnes2020umap} and MDS \cite{kruskal1964nonmetric} for our task, but found that UMAP did not determine sufficiently separable clusters and MDS was dropped due to its memory complexity of $O(N^2)$. 
Figure \ref{fig:cluster_w_sample} shows a visualisation of the resulting embedding and compares SNR profiles within the same cluster with those of another cluster.
On top of this embedding, we use DBSCAN \cite{ester1996density} to automatically identify clusters based on the low-dimensional representations. 
Since data points of a single cluster share similar SNR characteristics, we assume that connections assigned to the same cluster have the same connection state. 

\subsection{Visualization and anomaly discovery}
Based on cluster labels from \ref{sec:node_state} we transform a series of SNR profiles into a series of connection states. 
Using this series, we propose a connection state diagram to keep track of single PLC connections.
Therefore, we exploit the natural aesthetic of radial visualizations for periodic diagrams \cite{draper2009survey}, to visualize the state of a PLC connection.
The main contribution of this radial visualisation scheme is an intuitive relation to time, the ability to easily distinguish between noise and trend, and the possibility to spot events like fuse failures. 
Figure \ref{fig:data_overview} shows an example.

\begin{figure*}[h!]
    \centering
    \includegraphics[width=\textwidth,trim={0cm 0cm 0cm 0cm},clip]{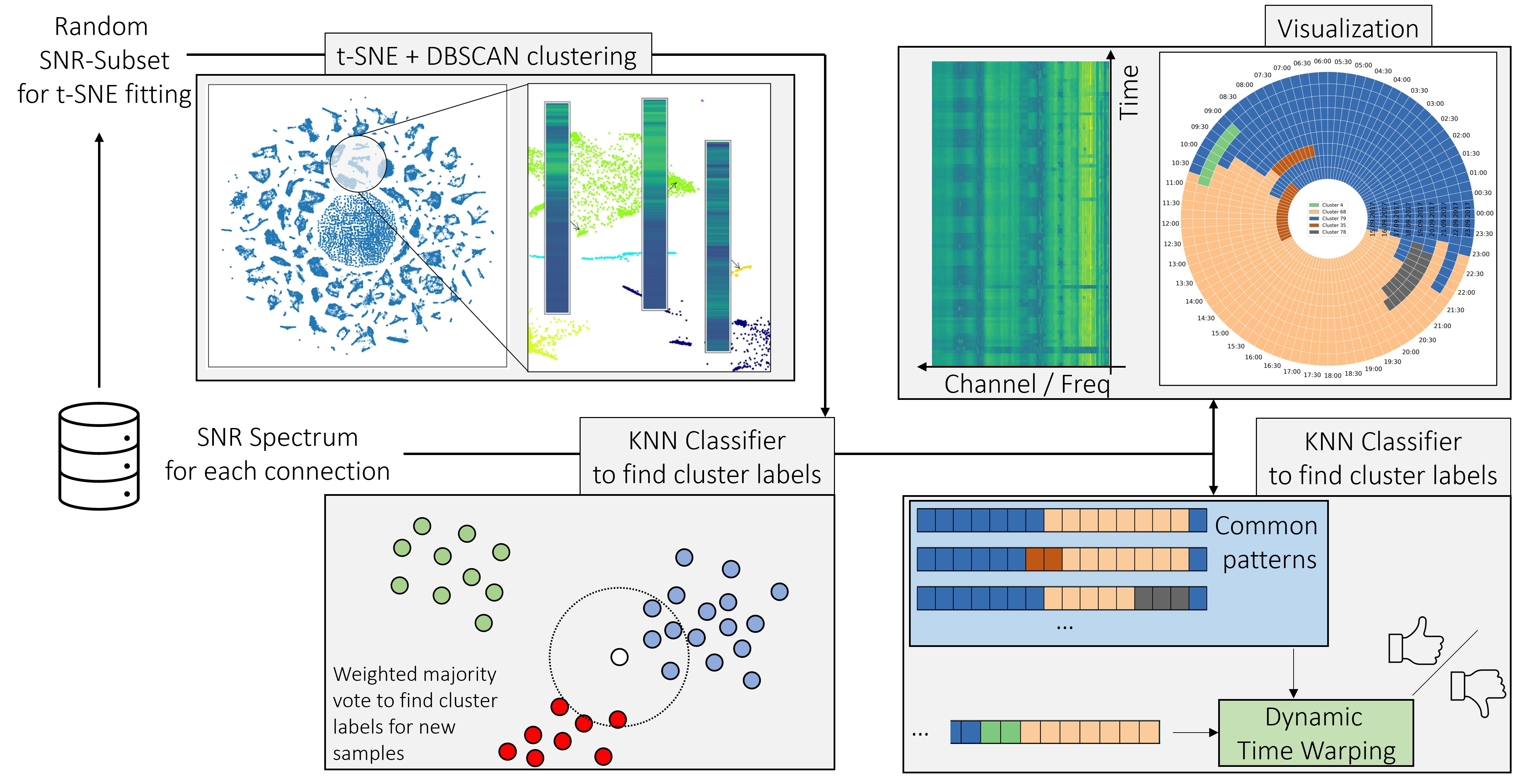}
    \caption{The overall pipeline of our grid monitoring system. Starting from a subset of FiN, we apply t-SNE to create a low-dimensional representation that serves as input for DBSCAN. Using a k-nearest neighbour (KNN) search, we assign all other SNR measurements to one of the clusters found by DBSCAN. Based on these cluster assignments, which we refer to as connection states, we transform SNR sequences into connection state sequences. Subsequently, we attach a radial visualization scheme for manual inspection and provide a dynamic time-warping system to detect anomalous connection states.}
    \label{fig:anomaly_discovery}
\end{figure*}

To demonstrate the use of connection states for the detection of anomalies, we propose a system based  on Dynamic Time Warping.
We used Dynamic Time Warping (DTW \cite{muller2007dynamic}) to find typical connection state patterns that occur during the normal daily operation of a node, as well as abnormal behaviour. 
Therefore, based on time, we split the FiN-2 dataset up into two parts. 
While the first 75\% of each temporal sequence of connection states are used for training, the later 25\% are used as an evaluation sequence which potentially shows anomalies. 
During training, DTW finds common patterns (templates) that occur during normal operation of each node. 
Small temporal changes to these patterns can be ignored during pattern matching by using a threshold which is a hyperparameter. 
Moreover, DTW can incorporate different metrics like euclidean or cosine to achieve different behaviour during the anomaly detection phase. 
Figure \ref{fig:anomaly_discovery} shows the complete architecture of our system and Figure \ref{fig:dtw} illustrates the results of our DTW anomaly detection system.

\subsection{Cable joints detection}
A typical procedure to investigate the influence of electrical components, such as cable joints, on the PLC or SNR signal would be a comprehensive laboratory study.
However, there can be large differences between laboratory measurements and the real circumstances in practical use.
While the environmental influences in the laboratory are largely known, the influences on the SNR spectrum under real use cannot be outlined in detail. 
There is an unknown number of interference sources, such as poorly shielded electronics, as well as a large variance of installed cable lengths, cable types, cable joints, quality of installation, consumers, etc.
Even switching operations of street lighting can have an influence on the SNR spectrum.
We propose an evaluation of electrical components driven by ML and data from practical application.
For that, we used the FiN-2 dataset, since it holds a comprehensive collection of metadata related to cable properties. 
Based on the SNR spectrum, we trained a ResNet18 \cite{he2016deep} to predict the number of connection cable joints installed within a cable section using the mean average error.
Therefore, we presented a spectrum of 96 SNR measurements, corresponding to a whole day, to the model. 
Furthermore, we use a supervised contrastive loss \cite{khosla2020supervised} to improve our results and perform a regression activation mapping \cite{selvaraju2017grad} to highlight regions that show a high sensitivity with respect to the regression result. 
Overall, our model was able to achieve a mean absolute error of $0.82$ on the validation set of this task.

\begin{figure*}[h!]
    \centering
    \includegraphics[width=\textwidth,trim={0cm 0cm 0cm 0cm},clip]{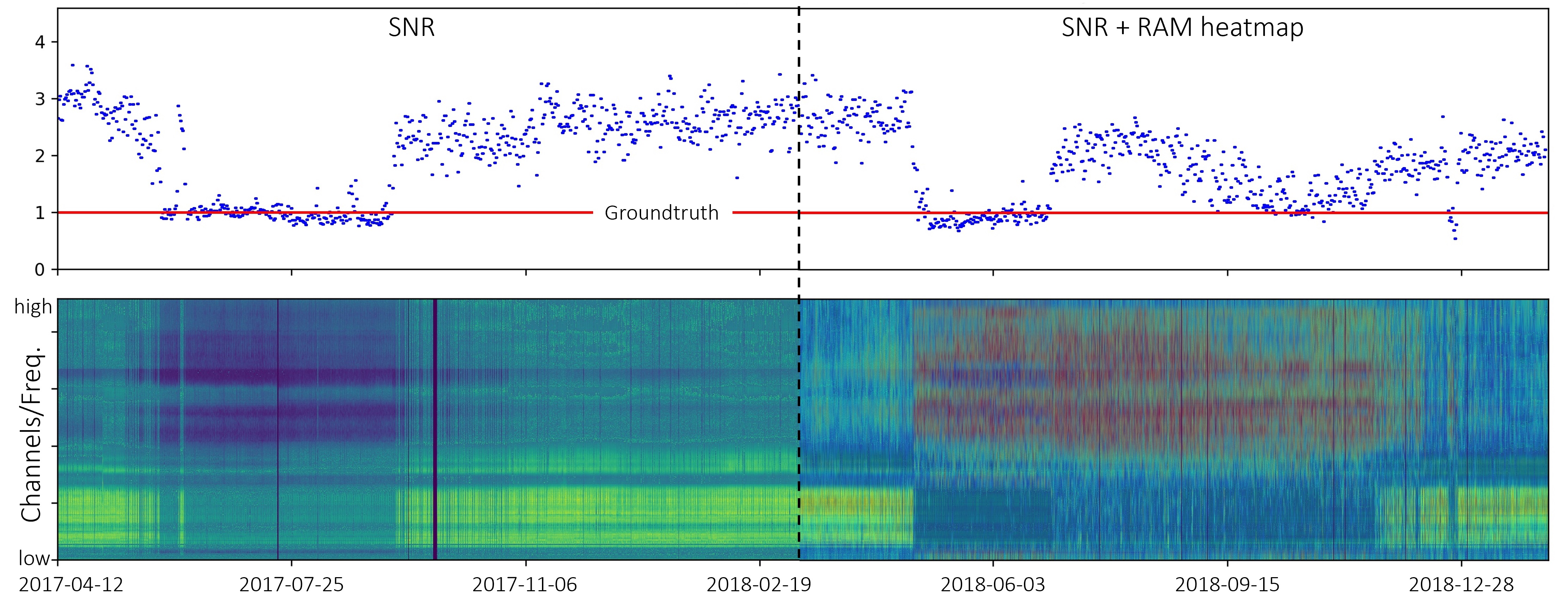}
    \caption{Results of our cable connection detection system for one connection. The regression results of our systems (top) for the corresponding SNR section (bottom). The left side shows the raw SNR spectrum, while the right side shows its continuation but with an additional overlay generated by a regression activation mapping. The regression task is performed on a section of the SNR spectrum representing a whole day. Sections with reliable estimation often show distinctly different activation patterns in the overlay.}
    \label{fig:sleeve_plot}
\end{figure*}

\begin{figure}[h!]
    \centering
    \includegraphics[width=0.5\textwidth,trim={0.1cm 0cm 2cm 0cm},clip]{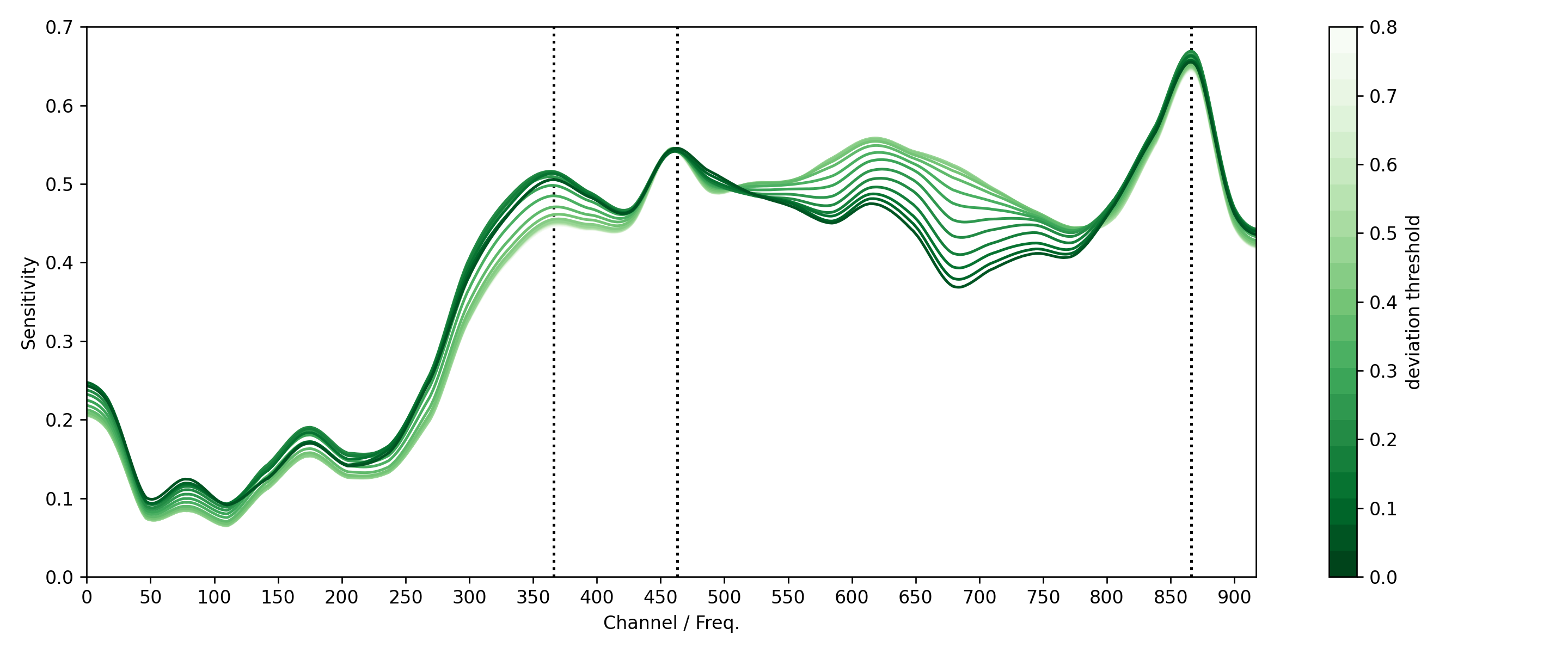}
    \caption{Mean sensitivity per channel when estimating cable joints. When estimating the number of cable joints within a cable section, each channel has a different mean sensitivity when considering the entire validation data. Looking only at the results that show a deviation to a minimum threshold (dark green), three peaks become visible. These three peaks correspond to the channels that seem most relevant for a correct estimation of the joints and reflect a characteristic of the joints. In this case, the connection joints were investigated.}
    \label{fig:sleeve_sensitivity}
\end{figure}

A closer analysis of the results shows that the precision of the prediction is strongly distorted by noise in some stages. 
During these periods with strong noise components, the predictions fluctuate significantly, whereas at other moments the prediction is very precise.
As mentioned before, this behaviour is in line with our expectations, since noise components are considerably larger compared to the SNR characteristic of a cable joint.
Using regression activation mapping, it can be observed that the regions used for prediction differ significantly in these areas of the SNR spectrum.
Figure \ref{fig:sleeve_plot} shows an example of this.
Beyond the prediction of the number of cable joints, however, we also want to draw conclusions about the transmission properties of the joints. 
For this purpose, based on the validation data, the portions of the SNR spectra on which a reliable prediction of the cable joints was possible are determined and illustrated in Figure \ref{fig:sleeve_sensitivity}. 
For these, the regression activation mapping is evaluated in detail and a distribution over the frequencies is calculated.
The distribution shows which frequencies of the SNR spectrum are associated with a high gradient and thus a high sensitivity with respect to the regression result.
During the comparison of all distributions, it is noticeable that in the regions where the prediction of the number of cable joints worked reliably, similar frequencies were always dominant.
This indicates that these are frequencies that are characteristically influenced by installed joints.
By using ML, it is thus possible to evaluate relevant frequency characteristics of joints during practical use.
In addition, insights gained into how joints affect the SNR spectrum can be used to refine model predictions. 

\subsection{PLC network topology estimation}
A major benefit from using PLC is that nodes communicate in a broadcast-fashion. 
This allows the PLC network, on the one hand, to establish communication links between nodes that are multiple hops away, which keeps the overall hop count for communication small, but on the other hand, it hides the actual PLC network topology.
Since it is not possible to distinguish between these two types of neighbours, it is necessary to record the data of all connections for complete grid monitoring. 
This means connections to real neighbours as well as to indirect neighbours, to which there is only an indirect connection.
However, since the number of indirect neighbours is several times higher than the number of direct neighbours, this leads to a significant increase in storage for redundant data.
For an efficient storage of the data, it is necessary to filter out indirect connections. 
One more advantage of what is that the structure of the electricity grid can also be derived efficiently from the topology of the PLC network.
Subsequently, a clean PLC network topology opens a simple way to automatically create a digital twin of the lv electricity grid.\\

Different works that aim at predicting the actual PLC network topology show good performance in a laboratory setting. However, they fail when applied to realistic scenarios \cite{ahmed2013power}.
One reason for this is that such approaches often get tailored to simulated data due to the absence of real data or require expensive two-sided measurements in the field. 
Furthermore, the required set of measurements or constraints that are necessary to determine the PLC network topology can be a reason why an approach is not applied in practical use.
To overcome these limitations, we show how ML can be used to untangle the PLC signal and retrieve the actual PLC network topology.

\begin{figure}[h!]
    \centering
    \includegraphics[width=0.45\textwidth,trim={0cm 0cm 0cm 0cm},clip]{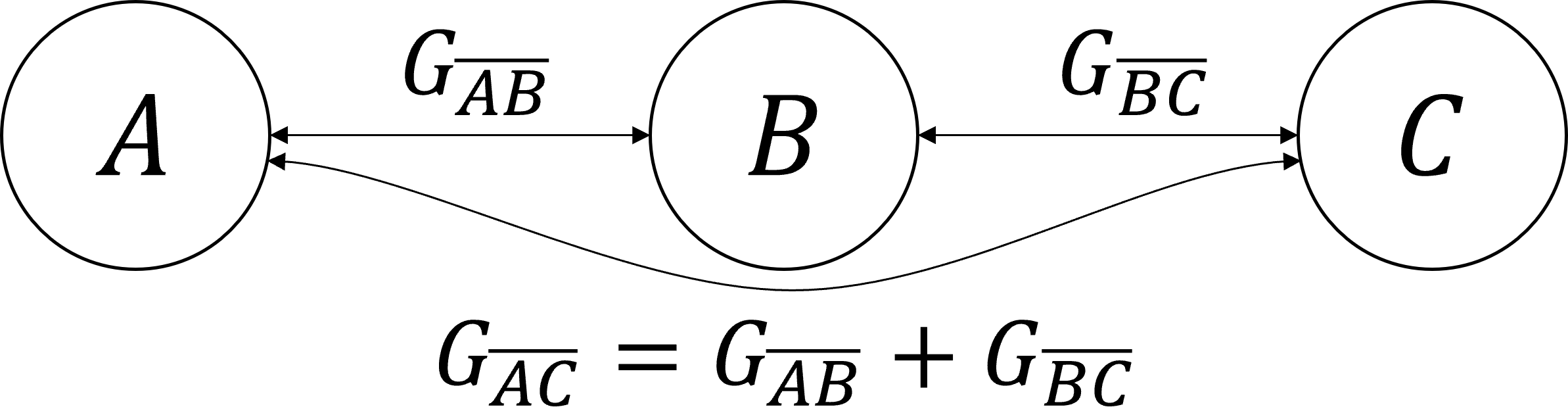}
    \caption{If the pattern, that is given by the transfer function of cable section $\overline{AB}$ to the SNR spectrum, can also be found in the SNR spectrum of cable section $\overline{AC}$, we can assume $C$ to be not directly connected to A. }
    \label{fig:transfer_function}
\end{figure}

\begin{figure*}[h!]
    \centering
    \includegraphics[width=1.0\textwidth,trim={0cm 0cm 0cm 0cm},clip]{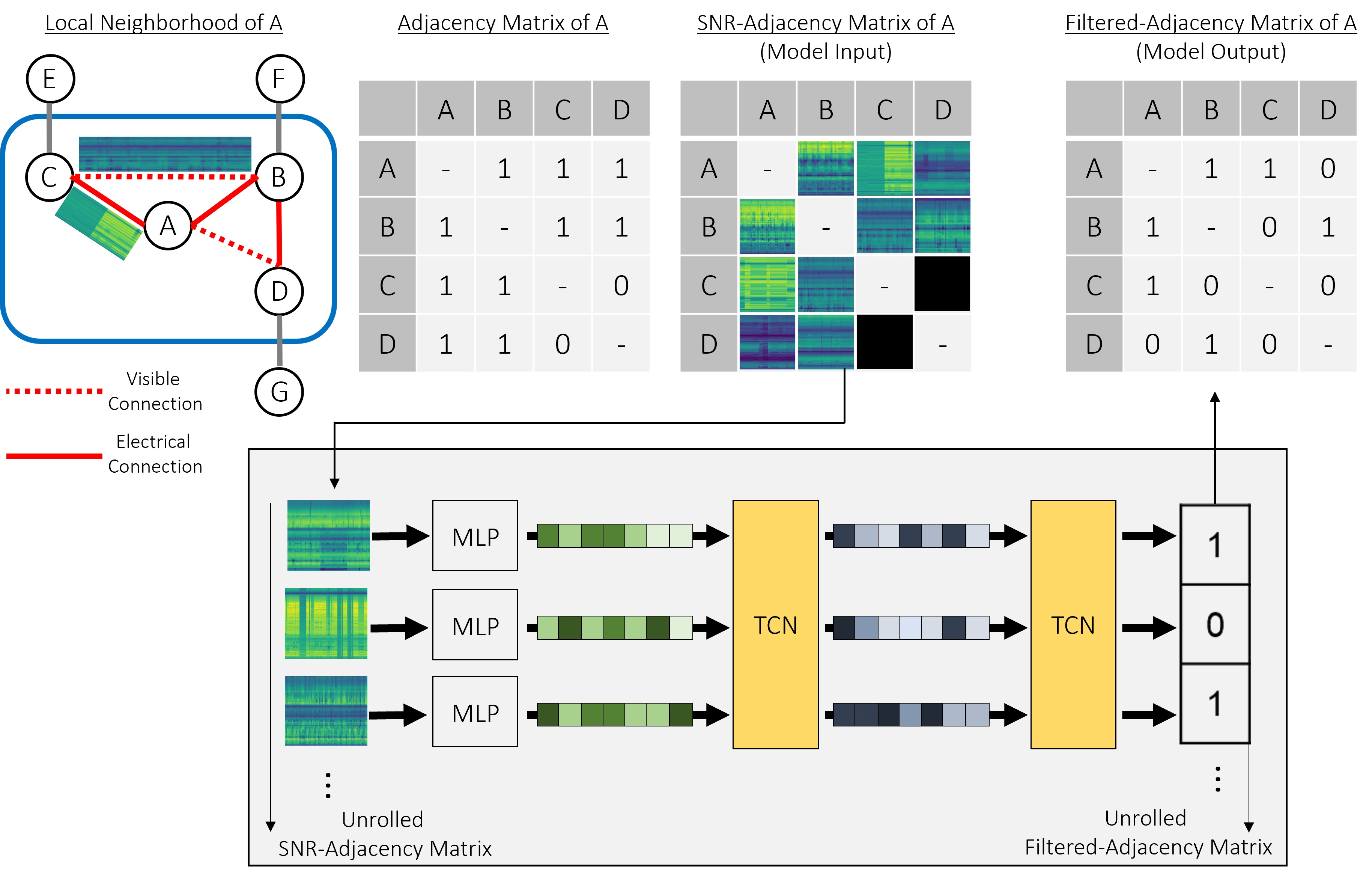}
    \caption{Overview of the neighbourhood estimation system. A local neighbourhood of a node within the electricity grid is used to create an adjacency matrix. The adjacency matrix shows both direct and indirect neighbours that may be several hops away. Enriched with the SNR profile of a single time step, the SNR adjacency matrix is passed to the ML model. A three-layer MLP is used as a feature encoder, while two TCN layers compute a filtered adjacency matrix showing only direct neighbours.}
    \label{fig:grid_tomography_arch}
\end{figure*}

\begin{figure*}[!ht]
    \centering
    \includegraphics[width=.75\textwidth,trim={0cm 0cm 0cm 0cm},clip]{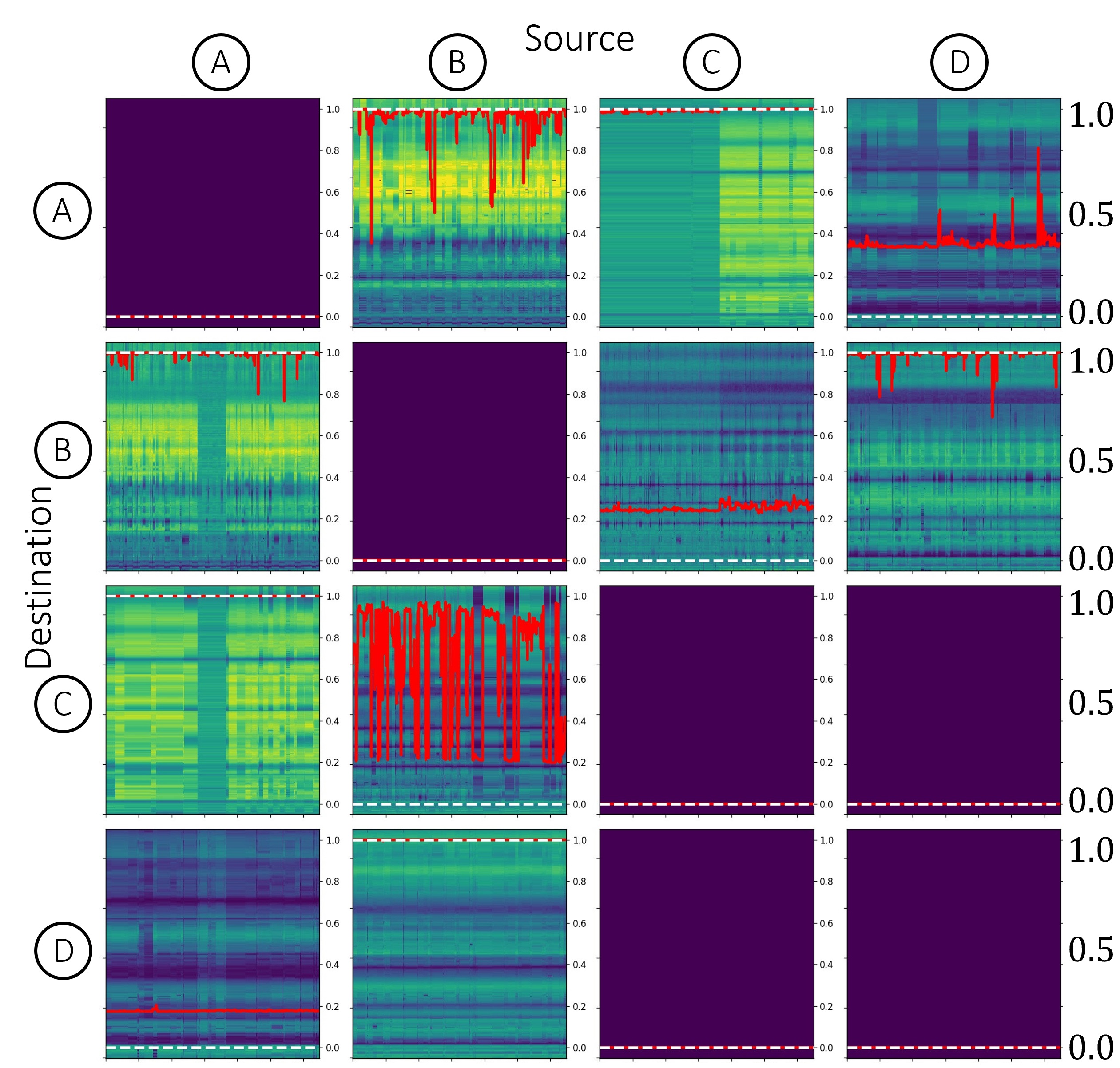}
    \caption{Result of the neighbourhood estimation system for a local neighbourhood. The neighbourhood is given in the form of the SNR adjacency matrix, while the ground truth of whether a neighbour is directly or indirectly connected is shown in white. The results of the system for the entire SNR sequence are shown in red. Since the SNR is measured in both directions, the matrix is approximately symmetric in theory. In practice, however, there are sometimes significant differences in the outward and return directions of the SNR spectrum. As the results can also vary in both directions, as can be seen on connection $\overline{BC}$, it is possible to further improve the results in post-processing due to this symmetry constraint.}
    \label{fig:grid_tomography_result}
\end{figure*}

Figure \ref{fig:grid_tomography_arch} shows the overall architecture of our system. 
Starting from a node whose neighbourhood is to be untangled, we use the SNR profiles of all neighbours as input to our ML model. 
We redefine the task of determining the PLC network topology to a filtering of local subgraphs. Therefore, we assume a local neighbourhood, around a specific node, to be a subgraph that can also be described by an adjacency matrix. 
Figure \ref{fig:grid_tomography_result} shows an example of such an adjacency matrix.
A neural network architecture, as shown in Figure \ref{fig:grid_tomography_arch}, is used to detect the presence of the transfer function of a given cable section in the SNR spectrum of another cable section. 
Thereby, we try to filter out PLC connections that show the fingerprint of multiple transfer functions and subsequently have to be multiple hops away.
As Figure \ref{fig:transfer_function} shows, having multiple fingerprints is an indicator that a node is an indirect neighbour.


For this task, we structure the SNR measurements of a single timestamp of a local neighbourhood in the form of the adjacency matrix.
We get an input size of $n \times n \times f$, where $n$ is the number of nodes in the local neighbourhood and $f$ is the number of covered channels in the SNR spectrum, here 917.
Since we assume that the fingerprint of the cable connections in the SNR is static in a limited time range, we only process single time steps in our model. 
We use a three-layer multilayer perceptron (MLP) in an unsupervised autoencoder to find good preinitialised weights for the network.
After this pretraining phase, we incorporate the MLP as feature encoder for the SNR data. 
Building up on those features, we use a Temporal Convolutional Network (TCN \cite{oord2016wavenet}) to identify a clean adjacency matrix of the local neighbourhood.
This clean adjacency matrix is now up to show only connections that have a real wired connection.
In advance, TCNs are unbound to a specific input length and allow the model to deal with local neighbourhoods of different size.
Figure \ref{fig:grid_tomography_arch} shows the inputs and results of our model, which was able to achieve an accuracy of $92\%$ at the level of individual entries in the adjacency matrix and $55\%$ at the level of the full matrix.
Building on top of these raw results, additional heuristics could be incorporated to further improve the results. 
Since the SNR is measured in both directions, some incorrect results can be corrected by looking at them together.
Moreover, the local neighbourhoods can be considered overlapping, so that a single connection can be checked in different neighbourhoods and further errors can be eliminated.
Furthermore, it is also possible to include other data, such as voltage. 
However, since we only provide this application as an example, we leave these extensions open for further research. 

\section{Conclusion}
In this paper, we publish the first PLC dataset of its kind. This dataset stands out for the large number of PLC nodes as well as for the large amount of metadata it includes. 
This dataset, called FiN-2, was recorded during real deployment in the electricity grid at lv-level. 
Thus, for the first time, it is possible to investigate the broad spectrum of influences on the SNR spectrum of a PLC network, as well as the possibilities of broad monitoring of the voltage at lv-level using real data.
To demonstrate the utility of this data, we have successfully implemented a number of ML-based example applications.
We have demonstrated how a large-scale deployment of PLC nodes allows wide-area monitoring, even at lv-level. 
Building on this, we have shown that it is possible to group the large number of different SNR profiles into clusters that can be used as connection state.
Unexpected deviations from this state build the basis to detect anomalies.
Apart from that, we could show that it is possible to deeply investigate properties of a cable section by ML. 
For this purpose, we identified the number of installed joints solely based on the SNR spectrum and drew conclusions on the influence of a joint on the SNR spectrum.
Finally, we were able to show that it is possible to detect the topology of a PLC network utilizing the SNR spectrum and to draw conclusions about the electrical network.

\section{Methods}
\subsection{Background – PLC}\label{sec:plc_background}
Broadband powerline communication (PLC) is a way of setting up information and communications technology (ICT) in the electricity grid, especially at the low-voltage level (lv-level), in a relatively simple and cost-effective manner.
Historically, this equipment was not necessary at lv-level.
However, the advancing energy transition and the associated increase in the number of distributed energy resources is necessary, for example, to read out intelligent metering systems or to control decentralized power generation systems and regulate them as required.
PLC operates according to the IEEE 1901 standard and uses the power cable as a transmission path for high-frequency signals in a frequency range between 2 and 28 MHz, which makes this technology cheaper than others.
Communication is based on orthogonal frequency division multiplexing, in which the entire frequency range is divided among individual carrier frequencies with a bandwidth of approx. 24.414 kHz.
917 of the carrier frequencies can be used for communication.
Some frequencies are protected for other applications and may not be used for communication.

In addition to this primary benefit of PLC, there is a secondary benefit that is at least as important: live grid monitoring.
Apart from voltage, PLC modems also measure the signal-to-noise ratio (SNR), which is specified in decibels (dB).
The SNR indicates how high the power of the received signal $P_{RX}$ is in relation to the power of the noise $P_{N}$ \cite{ferreira2011power}.
The received power depends on the transmit power $P_{TX}$ and the complex transfer function $\underline{H}$.
All these quantities are time and frequency dependent.
The transfer function contains information about the transmission path, i.e. essentially about the power cable and the topology of the grid.

\section{Data availability}
FiN-1 dataset is available under Creative Commons CC BY 4.0 International and can be downloaded from Zenodo \cite{FiNDataset}. The FiN-2 dataset is currently being revised for publication and will be published also under Creative Commons CC BY 4.0 International \cite{FiN2Dataset}.

\appendices
\section*{Acknowledgment}
The presented work in this publication is based on research activities, supported by the Federal Ministry of Education and Research (BMBF) and the Projektträger Jülich (PTJ), the described topics are included in the project “Fühler im Netz 2.0” (reference number: 03EK3540B).
Only the authors are responsible for the content of this paper.

\begin{figure}[!h]
   \begin{minipage}[!h]{.3\linewidth} 
      \includegraphics[width=.8\linewidth]{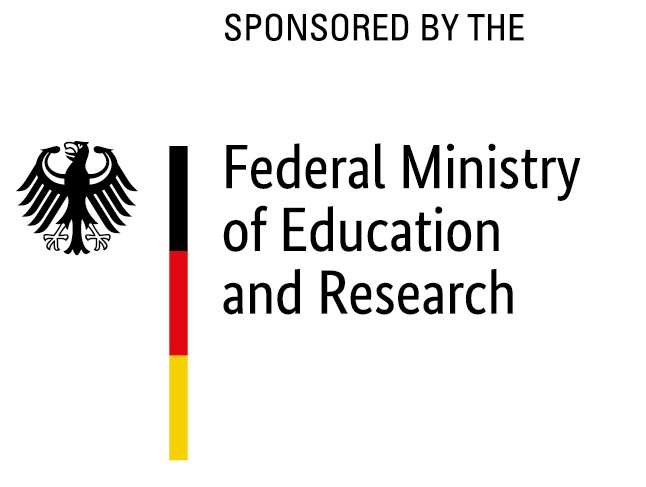}
   \end{minipage}
   \hspace{.1\linewidth}
   \begin{minipage}[!h]{.3\linewidth} 
      \includegraphics[width=.7\linewidth]{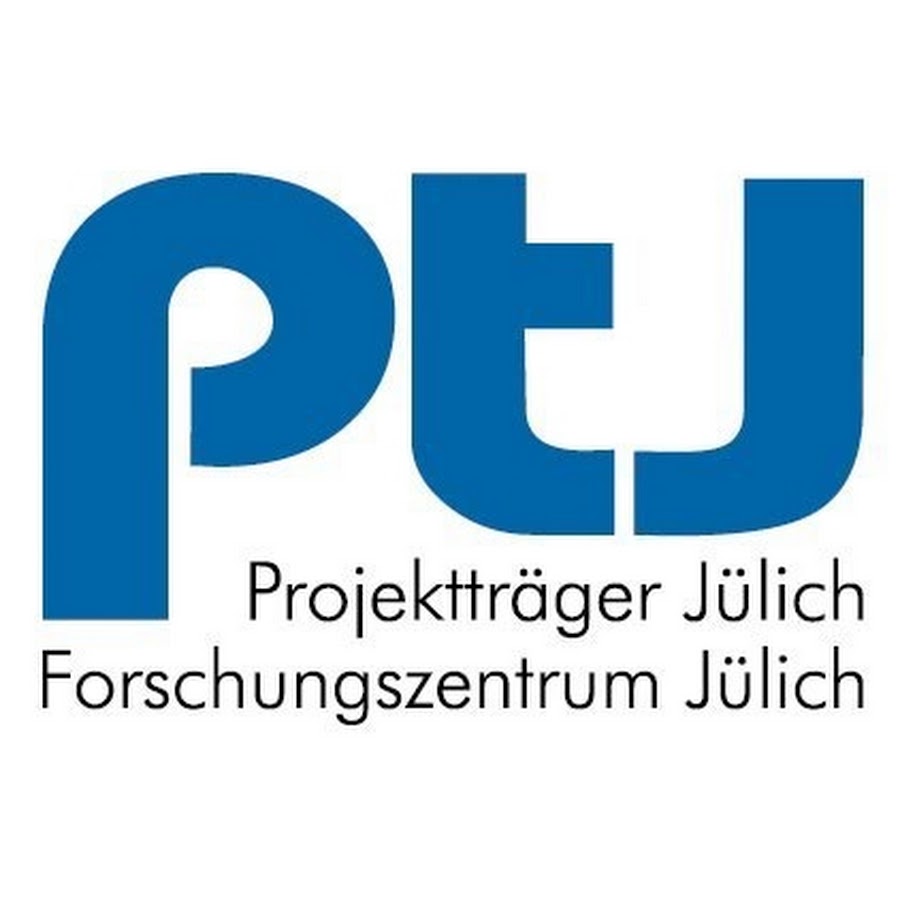}
   \end{minipage}
\end{figure}

\ifCLASSOPTIONcaptionsoff
  \newpage
\fi



\bibliographystyle{IEEEtran}
\bibliography{bibtex/bib/bibliography}
%

%

\begin{IEEEbiographynophoto}{Christoph Balada}
received the B.Sc. and M.Sc. in Computer Science and Systems Engineering from the University of Ilmenau, Germany in 2019. He is currently a research associate at the German Research Center for Artificial Intelligence (DFKI GmbH), Kaiserslautern, where he is also pursuing a PhD in machine learning. His current research interests include machine learning in general and the application of Convolutional and Graph Neural Networks in the context of Smart Grids. 
\end{IEEEbiographynophoto}

\begin{IEEEbiographynophoto}{Sheraz Ahmed}
received the master’s degree in computer science from Technische Universitaet Kaiserslautern, Germany, and the Ph.D. degree from the German Research Center for Artificial Intelligence, Germany, under the supervision of Prof. Dr. Prof. H. C. A. Dengel and Prof. Dr. habil. M. Liwicki. His Ph.D. topic was Generic Methods for Information Segmentation in Document Images. Over the last few years, he has primarily worked on the development of various systems for information segmentation in document images. From 2012 to 2013, he visited Osaka Prefecture University, Osaka, Japan, as a Research Fellow, supported by the Japanese Society for the Promotion of Science, and in 2014, he visited The University of Western Australia, Perth, Australia, as a Research Fellow, supported by the DAAD, Germany, and Go8, Australia. He is currently a Senior Researcher with the German Research Center for Artificial Intelligence (DFKI GmbH), Kaiserslautern, where he is leading the area of time-series analysis. His research interests include document understanding, generic segmentation framework for documents, gesture recognition, pattern recognition, data mining, anomaly detection, and natural language processing. He has more than 30 publications on the said and related topics, including three journal papers and two book chapters. He is a frequent Reviewer of various journals and conferences, including Patter Recognition Letters, Neural Computing and Applications, IJDAR, ICDAR, ICFHR, and DAS.
\end{IEEEbiographynophoto}

\begin{IEEEbiographynophoto}{Andreas Dengel}
received the Diploma degree in CS from the University of Kaiserslautern and the Ph.D. degree from the University of Stuttgart. He was with IBM, Siemens, and Xerox Parc. In 1993, he became a Professor with the Computer Science Department, University of Kaiserslautern, where he holds the chair position in knowledge-based systems. Since 2009, he has been an appointed Professor (Kyakuin) with the Department of Computer Science and Information Systems, Osaka Prefecture University. He is currently the Scientific Director of the German Research Center for Artificial Intelligence (DFKI GmbH), Kaiserslautern. Moreover, he is a Co-Editor of international computer science journals and has written or edited 12 books. He has authored more than 300 peer-reviewed scientific publications and has supervised more than 170 Ph.D. and master’s theses. His main scientific emphasises are on pattern recognition, document understanding, information retrieval, multimedia mining, semantic technologies, and social media. He is a member of several international advisory boards, has chaired major international conferences, and founded several
successful start-up companies. He is an IAPR Fellow and received prominent international awards.
\end{IEEEbiographynophoto}

\begin{IEEEbiographynophoto}{Max Bondorf}
received the B.Sc. and M.Sc. in electrical engineering from the University of Wuppertal, Germany in 2019.
He is currently a research associate at the Chair of Power Systems Engineering at the University of Wuppertal, where he is also pursuing his Ph.D. degree.
His current research interests include the benefits of broadband power line communication for smart grids and asset management.
\end{IEEEbiographynophoto}

\begin{IEEEbiographynophoto}{Markus Zdrallek}
studied electrical engineering at the Technical University of Darmstadt. He received his diploma degree in 1996. After working as a research assistant at the University of Siegen he received his Ph.D. degree in 2001. In the years 2000–2010 he worked in several leadership positions in the grid division of the Energy Supply Company RWE. Since 2010 he holds the chair for Power System Engineering at the University of Wuppertal.
\end{IEEEbiographynophoto}




\end{document}